\documentclass{article}

\usepackage[preprint, nonatbib]{neurips_2021}

\usepackage[utf8]{inputenc} 
\usepackage[T1]{fontenc}    
\usepackage{hyperref}       
\usepackage{url}            
\usepackage{booktabs}       
\usepackage{amsfonts}       
\usepackage{nicefrac}       
\usepackage{microtype}      
\usepackage{xcolor}         
\usepackage{graphicx}
\usepackage[ruled, vlined]{algorithm2e}
\usepackage[normalem]{ulem}

\newif\ifturnoffcomments
\turnoffcommentsfalse

\title{AutoDIME: Automatic Design of Interesting Multi-Agent Environments}

\author{%
  Ingmar Kanitscheider \\
  OpenAI \\
  \texttt{ingmar@openai.com} \\
  \And
  Harri Edwards \\
  OpenAI \\
  \texttt{harri@openai.com}
}

\begin{document}

\maketitle

\begin{abstract}
  Designing a distribution of environments in which RL agents can learn interesting and useful skills is a challenging and poorly understood task, for multi-agent environments the difficulties are only exacerbated. One approach is to train a second RL agent, called a teacher, who samples environments that are conducive for the learning of student agents. However, most previous proposals for teacher rewards do not generalize straightforwardly to the multi-agent setting. We examine a set of intrinsic teacher rewards derived from prediction problems that can be applied in multi-agent settings and evaluate them in Mujoco tasks such as multi-agent Hide and Seek \cite{baker2020emergent} as well as a diagnostic single-agent maze task. Of the intrinsic rewards considered we found value disagreement to be most consistent across tasks, leading to faster and more reliable emergence of advanced skills in Hide and Seek and the maze task. Another candidate intrinsic reward considered, value prediction error, also worked well in Hide and Seek but was susceptible to noisy-TV style distractions in stochastic environments. Policy disagreement performed well in the maze task but did not speed up learning in Hide and Seek. Our results suggest that intrinsic teacher rewards, and in particular value disagreement, are a promising approach for automating both single and multi-agent environment design.
  
\end{abstract}

\section{Introduction}

    In natural evolution, the simple yet powerful process of natural selection has led to the vast complexity of the living world. It is believed that one of the main drivers of biological complexity are evolutionary arms races between competing organisms \cite{dawkins1979arms}. A similar dynamic of competing agents creating new tasks for each other has been exploited in competitive multi-agent reinforcement learning (RL) to reach super-human performance in games such as Backgammon \cite{tesauro1995temporal}, Go \cite{silver2017mastering}, Dota \cite{openai2019dota} and Starcraft \cite{vinyals2019grandmaster} and to showcase impressive emergent complexity from simple game rules in physically grounded environments \cite{sims1994evolving, bansal2017emergent, jaderberg2019human, liu2019competitive, baker2020emergent}. Yet just as evolution happened in bursts at specific times and places \cite{uyeda2011million}, finding the precise environmental conditions that lead to interesting emergent skills in complex environments is often a time-consuming and laborious process. One strategy to mitigate this problem is to train agents on a wide distribution of randomized environments, in the hope that some small fraction of them will lead to the emergence of a new skill \cite{baker2020emergent, openai2019dota, jaderberg2019human}. 
    
    The technique of training agents on wide distributions of environments to obtain better and more robust skills, called {\it domain randomization}, has also been used in single-agent settings \cite{jakobi1997evolutionary, sadeghi2017sim2real, tobin2017domain, chebotar2018closing, openAI2019solving}. A growing number of recent works have designed automatic environment curricula that adapt the environment distribution over the course of training to maximize the fraction of learnable environments \cite{matiisen2019teacher, openAI2019solving, wang2019paired, wang2020enhanced, portelas2020teacher, campero2020learning, dennis2020emergent}.  

    A common approach to maximize learnability is to select environments that are neither too hard nor too easy based on a single-agent performance measure such as reward or success probability \cite{openAI2019solving, wang2019paired, campero2020learning, dennis2020emergent}. 
        
    The present paper generalizes curriculum learning for environment generation to multi-agent environments. We use the setup of Teacher-Student Curriculum Learning \cite{matiisen2019teacher, campero2020learning, dennis2020emergent}, where an RL-trained {\it teacher} samples environments of one or several {\it student} agents and is trained alongside the students. The teacher reward is chosen to incentivize the teacher to select environments that maximize student learning. A key challenge of competitive multi-agent environments is that the rewards or success probabilities of students are often not informative about student performance or learning progress: a student might achieve a high reward because they are highly skilled or because their opponents do badly and student rewards may fluctuate widely through skill discovery \cite{baker2020emergent}. 
    
    Instead, we focus on teacher rewards that evaluate students' return relative to some dynamic prediction. In particular, actor-critic algorithms such as PPO \cite{schulman2017proximal} train a value function to predict the return on each episode. A large difference between value prediction and value target (called {\it value prediction error}) might indicate an environment where students can learn something new \cite{jiang2020prioritized}. We also explore teacher rewards that measure the disagreement of an ensemble of student value functions with different initializations  ({\it value disagreement}) \cite{zhang2020automatic} and the disagreement between an agent's action distribution and the action distribution of a second, independently initialized policy that is trained to behaviorally clone the main again ({\it policy disagreement}). Large disagreement signals that a student is uncertain about the return or action distribution of an environment, which suggests that something new can be learned. Once a student has stopped learning in an environment, either disagreement measure is expected to converge to zero. 
    
    We also evaluate whether teacher rewards are susceptible to uncontrolled stochasticity in the environment. Just as intrinsically rewarded RL agents are sometimes attracted to ``noisy TV'' states with unpredictable transitions \cite{schmidhuber1991curious, oudeyer2007intrinsic, burda2018large}, a teacher may be incentivized to sample environments with unpredictable student returns, without any possibility of learning progress. We expect this problem to become more prevalent in more complex and harder-to-predict environments.
    
    In summary, our contributions are as follows:
    \begin{itemize}
        \item We show that intrinsic teacher rewards that compare student reward or behavior relative to some prediction can lead to faster skill emergence in multi-agent Hide and Seek and faster student learning in a single agent random maze environment.
        \item We formulate an analogue of the noisy TV problem for automatic environment design and analyze the susceptibility of intrinsic teacher rewards to uncontrolled stochasticity in a single agent random-maze environment. We find that value prediction error and to a small extent policy disagreement is susceptible to stochasticity while value disagreement is much more robust.
    \end{itemize}

\section{Related work}
{\bf Environment distributions for RL} Previous works in multi-task RL \cite{beattie2016deepmind, hausman2018learning, yu2020meta}, multi-goal RL \cite{kaelbling1993learning, andrychowicz2017hindsight} and meta RL \cite{wang2016learning, duan2016rl} designed a fixed distributions of tasks or goals to increase generalization. In domain randomization \cite{jakobi1997evolutionary, sadeghi2017sim2real, tobin2017domain, openAI2019solving}, one defines distributions of environments to obtain more robust skills.

{\bf Self-play} Multi-agent self-play has been used both to achieve super-human performance on predefined games \cite{tesauro1995temporal, silver2017mastering, openai2019dota, vinyals2019grandmaster} as well as a tool to explore novel skills in a single environment or fixed distribution of environments \cite{sims1994evolving, srivastava2013first, bansal2017emergent, baker2020emergent, jaderberg2019human}. In asymmetric self-play \cite{sukhbaatar2017intrinsic, sukhbaatar2018learning, liu2019competitive, openai2021asymmetric}, goal-setting and goal-following agents compete to improve generalization among single-agent tasks.

{\bf Curriculum learning} Recently there has been a lot of work on using automated curricula to speed up exploration in single-agent tasks \cite{florensa2018automatic, matiisen2019teacher, portelas2020teacher, wang2019paired, wang2020enhanced, zhang2020automatic, campero2020learning, dennis2020emergent, gur2021adversarial, jiang2020prioritized}. Environments or goals may be generated as in this work using RL \cite{matiisen2019teacher, campero2020learning, dennis2020emergent, gur2021adversarial}, or using GANs \cite{florensa2018automatic}, evolutionary algorithms \cite{wang2019paired, wang2020enhanced} or Gaussian mixture models \cite{portelas2020teacher}. Selection criteria for environments include intermediate student performance \cite{florensa2018automatic, wang2019paired, campero2020learning}, learning progress \cite{matiisen2019teacher, portelas2020teacher} and regret \cite{dennis2020emergent, gur2021adversarial}, but neither of these selection criteria has been generalized to or tested in multi-agent environments. Initial-state value disagreement \cite{zhang2020automatic} and value prediction error \cite{jiang2020prioritized} has been previously used to rerank randomly sampled goals or environments.

{\bf Prediction-based Exploration} A number of works have used an agent's ability to predict the future to design exploration bonuses \cite{schmidhuber1991possibility, oudeyer2007intrinsic, stadie2015incentivizing, pathak2017curiosity, burda2018large, houthooft2016vime, choshen2018dora}. In \cite{pathak2019self, shyam2019model}, exploration bonuses were calculated from the disagreement between several prediction models. The term ``noisy TV'' was coined in \cite{burda2018large} based on previous observations \cite{schmidhuber1991curious, oudeyer2007intrinsic} that prediction errors due to stochasticity and model misspecification are not helpful for exploration. 

\section{Teacher-Student Curriculum Learning}
\label{sec:tscl}

Teacher-Student Curriculum Learning (TSCL) \cite{matiisen2019teacher} is a training scheme where an RL-trained teacher samples environments in which student agents are trained. The teacher is rewarded for generating environments where student can learn most according to some measure of student behavior. Domain randomization, where the environment distribution is not adapted during training, can be considered a special case of TSCL: If the teacher is trained using maximum-entropy RL with constant teacher reward the teacher policy will converge to a stationary distribution.

TSCL is a very general scheme that can be combined with any single- or multi-agent RL environment. In our setup, the teacher first samples an environment at the beginning of a student episode in a single time step, the student policies are then rolled out and the teacher reward is calculated at the end of the episode (see Appendix \ref{app_dime} for pseudo-code).

The teacher can either specify the environment fully (``joint sampling'') or partially (``conditional sampling''). Under joint sampling, all environment parameters $Z$ are determined by a sample of the teacher policy. Under conditional sampling, the environment parameters are split into $Z = (X, Y)$, such that $Y$ is sampled from a fixed distribution $p(Y)$ and then given as observation of the teacher. The role of the teacher is to only generate the remaining parameters $X$, i.e. to specify $p_\theta(X|Y)$. $X$ and $Y$ should be chosen such that for any $Y$ there are both easy and hard environments depending on the choice of $X$.

We find that conditional sampling has two advantages: First, it is often easier to implement, because the teacher does not need to interact with every random sampling step of a procedurally generated environment such as the random maze in section \ref{sec:random_maze}. Second, in the case of Hide and Seek (section \ref{sec:hide_and_seek}), we find empirically that having the teacher specify fewer environment parameters (only the spawn locations of boxes and ramps) leads to better performance than having it specify more environment parameters (the spawn locations of agents, boxes and ramps). We speculate that conditioning the teacher on a fixed sampling distribution acts as a type of domain randomization for the teacher that prevents the teacher policy from narrowing in on a too small subset of the distribution of currently learnable environments.

\section{Teacher rewards}
\label{sec:teacher_rewards}
   We consider teacher rewards that are domain-general and applicable to multi-agent environments. Desirable teacher rewards sample environments where students can make learning progress and do not oversample environments with more stochastic student returns over environments with less stochastic student returns (a variant of the noisy TV problem \cite{burda2018large})
   
   We assume that students are trained using an actor-critic algorithm such as PPO \cite{schulman2017proximal} where a value function critic $V(s_t)$ predicts future returns. In our implementation the target $\hat{V}(s_t)$ for the value function critic is implemented using General Advantage Estimation (GAE) \cite{schulman2015high}.
   
   We assume that all student episodes have equal length. The total teacher reward for each episode is obtained by summing one of the following per-timestep rewards over episode time, $r = \sum_t r_t$:
   \begin{itemize}
        \item {\it Value prediction error}: $r_t = |V(s_t) - \hat{V}(s_t)|$.  A high value prediction error might indicate an environment where student learning has not yet converged. However, as we will see, value prediction error might also be high in environments with large unpredictable stochasticity.
        \item {\it Value disagreement}: $r_t = \textrm{std}_i V_i(s_t)$, where $V_i(s_t)$ are independently initialized value functions that are trained with the same value target $\hat{V}(s_t)$. We use ensemble size 2, where $r_t = \frac{1}{2}|V_1(s_t) - V_2(s_t)|$. We expect value disagreement to be high for environments where students are still uncertain about the return. In an environment where student learning has converged we expect both value functions to converge to their expected value target and value disagreement to converge to zero. 
        \item {\it Policy disagreement}: We train a second policy $\pi_2(a|s)$ using behavioral cloning (i.e. by minimizing $KL(\pi_1||\pi_2)$ on the student rollouts). The teacher reward is also given by $r_t = KL(\pi_1(a_t|s_t)||\pi_2(a_t|s_t))$, i.e. it is adversarial to the cloned agent. We expect policy disagreement to be high for environments in which the cloned policy has not yet matched the action distribution of the main policy. As for value disagreement, policy disagreement should converge to zero in environments where the students have stopped learning. Note that unlike value disagreement, policy disagreement is only dependent on student behavior, but not the reward function of the environment.
    \end{itemize}

    For multi-agent environments, we also average the teacher reward over students. In Hide and Seek, we sample each teacher and student policy with 10\% probability from a past policy to prevent cycling. For the calculation of the teacher reward, we only average over students that are rolled out using the current policy.

\section{Evaluation}
A well-designed teacher should lead to faster student learning than baseline training with a uniform or stationary environment distribution. However, it may be misleading to evaluate the student's performance under the teacher-generated training distribution, because the student may just do well because the teacher selected easy environments. Instead we evaluate the student under a fixed environment distribution that is independent of the teacher. Our evaluation distribution in Hide and Seek is the uniform distribution, in the random maze task it is the uniform distribution over ``hard'' environments.

\section{Experiments}
All environments in our experiments are physics-based environments simulated using the MUJOCO engine \cite{todorov2012mujoco}. Agents are point-like objects that can move and rotate by applying a force. Agents have an additional pull action that allows them to drag objects. In Hide and Seek, agents can also lock and unlock a (subset of) objects in place.

Students' observations are as described in \cite{baker2020emergent}: They observe their own location, orientation and velocity and the location, orientation and velocities of objects and agents around them. Objects and agents that are hidden are masked out for the policy, but visible to the value function (the mask is only used in Hide and Seek, not in the maze environment). In addition, they have access to a 360-degree LIDAR-type observation that informs them about the distance to the nearest object. In addition to the observations in \cite{baker2020emergent} agents also observed episode time, which we find improves the value function fit for longer episode lengths. We utilize decentralized execution and centralized training for multi-agent environments.

The teacher observes a 30x30 occupancy map of the environment where walls and object types (boxes, ramps, hiders, seekers) are encoded using different channels. The teacher observation is then processed through a feed-forward IMPALA conv-net \cite{espeholt2018impala}. The teacher action specifies the $x$ and $y$ grid location of objects such as agents, boxes and ramps. All locations are sampled simultaneously using a factorized distribution. If the grid location specified by the teacher would cause a collision with a wall or object we change the location to a nearby location that avoids the collision.

Teacher and student policies are optimized with PPO \cite{schulman2017proximal} and GAE \cite{schulman2015high} using large-scale distributed RL. See Appendix \ref{app_policy} and \ref{app_hyper} for further policy and hyperparameter details.

\subsection{Hide and Seek quadrant environment}
\label{sec:hide_and_seek}

The environment is a slightly modified version of the Hide and Seek quadrant environment (see section A.3 of \cite{baker2020emergent}): A team of 2 hiders is tasked to avoid line of sight from 2 seekers, and the seekers are tasked with keeping vision of the hiders. Agents can grab boxes and a ramp; the boxes can be locked in place. Hiders are given a preparation time at the beginning of the episode where the seekers are immobilized, giving the hiders a chance to run away or change the environment. Agents are given team-based rewards (hiders are given a per-time step reward of 1 if all hiders are hidden and -1 if any hider is seen by a seeker; seekers are given the opposite reward of hiders).

\begin{figure}
    \centering
    \includegraphics[width=\linewidth]{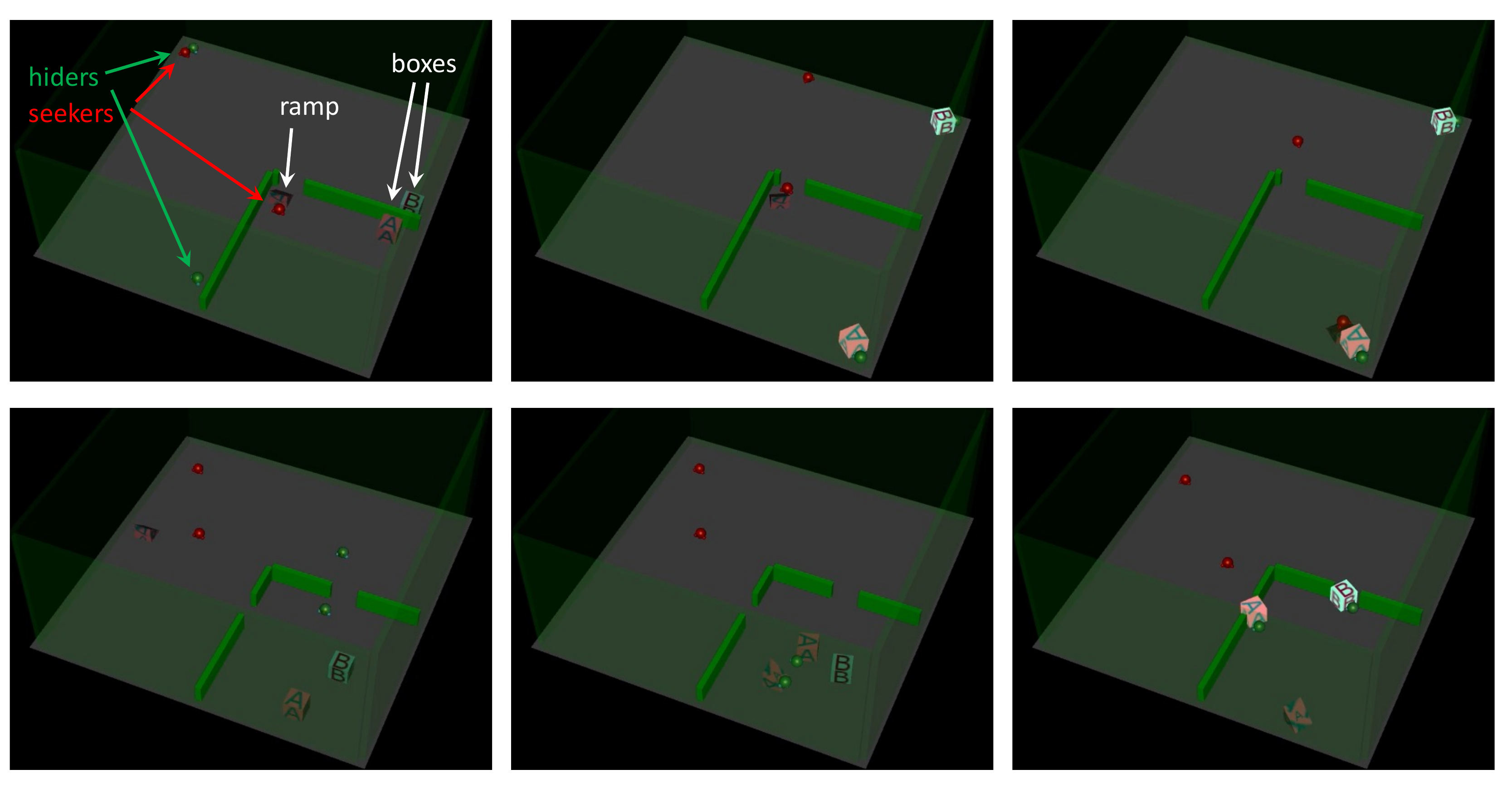}
    \caption{Hide and Seek quadrant environment, hiders are green and seekers red. {\it Top}: Example trial for phase 3 behavior: {\it Left}: Spawn locations of agents and objects, {\it Middle}: During the preparation time, hiders use boxes to barricade themselves in top and bottom right corners. {\it Right}: A seeker uses the ramp to climb the barricade of one of the hiders in the lower right corner, {\it Bottom}: Example trial for phase 4 behavior: {\it Left}: Spawn locations, {\it Middle}: During the preparation time, hiders drag the ramp inside the room, keeping it out of reach of seekers, {\it Right}: Hiders also barricade doors with boxes, thereby preventing seekers from entering the room or accessing the ramp after the preparation time has ended.}
    \label{fig:hs_snapshots}
\end{figure}

In the ``quadrant'' setup (Figure \ref{fig:hs_snapshots}), there is a room in the lower right corner that is connected by 1-2 doors at varying positions to the remainder of the environment. Hiders, seekers and the ramp are spawned randomly in the environment, but both boxes are only spawned inside the room. Both hiders and seekers receive a negative reward of -10 if they leave the environment or if they bring an object outside the environment.

The main change relative to \cite{baker2020emergent} is that we increase the size of the environment from 6m to 9m. A larger environment makes skill discovery more difficult, because skill discovery depends on agent discovering by chance that manipulating an object in a certain manner (such as seekers using the ramp to overcome obstacles or hiders taking the ramp out of reach of the seekers) gives them an advantage and the probability of such a chance manipulation is smaller in a larger environment. We also increase the episode length from 80 to 160 time steps to give agents more time to traverse a larger environment. 


In the standard conditional sampling setting, we sample an environment with random doors and random agent locations and supply the resulting occupancy map observation to the teacher. The teacher controls the spawn location of the ramp and both boxes and we use an action mask \cite{openai2019dota, vinyals2019grandmaster} to ensure that boxes are sampled close to the lower-right room, as under the uniform baseline. In the ``joint sampling`` setting, the teacher also controls the spawn locations of hiders and seekers (but this setting is still conditional in the sense that the teacher observes the number and locations of the randomly sampled doors).

The quadrant environment gives rise to the following skill progression: Basic running and chasing (phase 1; hider reward decreases), hiders use boxes to barricade themselves (phase 2; hider reward increases), seekers use ramps to overcome barricades (phase 3; hider reward decreases), hiders drag ramps inside barricades to prevent seekers from using them (phase 4; hider reward increases). Agents reach phase 1 and 2 very soon in all conditions, yet phase 3 and 4 is sometimes only reached with the help of a teacher. Typical snapshots of phase 3 and 4 behavior are shown in figure \ref{fig:hs_snapshots}.

\begin{figure}
    \centering
    \includegraphics[width=\linewidth]{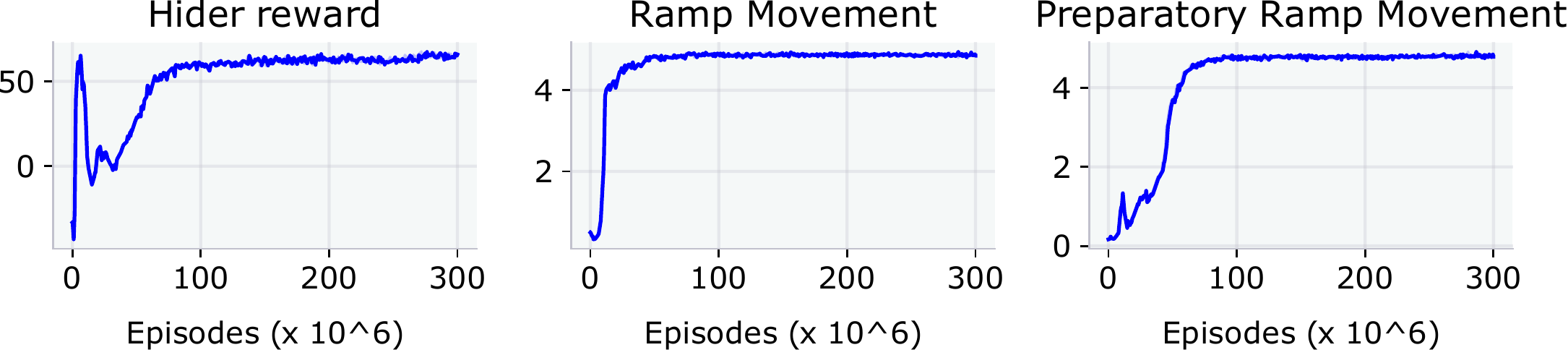}
    \caption{Skill progression through all 4 phases of Hide and Seek with value disagreement teacher reward under uniform evaluation (single seed): The first spike in hider reward ({\it left}) is caused by hiders barricading themselves with boxes (phase 2). Once seekers learn to overcome boxes with ramps (phase 3), hider reward decreases ({\it left}) and total ramp movement increases ({\it middle}). After around 40 million episodes hiders learn to drag the ramp away from the seekers during preparation time (phase 4). Both hider reward ({\it left}) and ramp movement during preparation time ({\it right}) increases. }
    \label{fig:hs_example}
\end{figure}

Skill progression cannot be measured directly using hider (or seeker) reward, since the reward fluctuates heavily through skill progression (Figure \ref{fig:hs_example}, left). Instead, emergent skills can be mapped to behavioral shifts in the way agents interact with the tools in their environment \cite{baker2020emergent}. Ramp use by seekers (phase 3) can be detected by an increase in the average displacement of the ramp (Figure \ref{fig:hs_example}, middle), because seekers need to drag the ramp to wherever hiders have barricaded themselves. Ramp defense by hiders (phase 4) can be detected by an increase in the average ramp displacement during preparation time, because hiders need to drag the ramp out of the reach of seekers as long as seekers are immobilized (Figure \ref{fig:hs_example}, left).

\begin{figure}
    \centering
    \includegraphics[width=\linewidth]{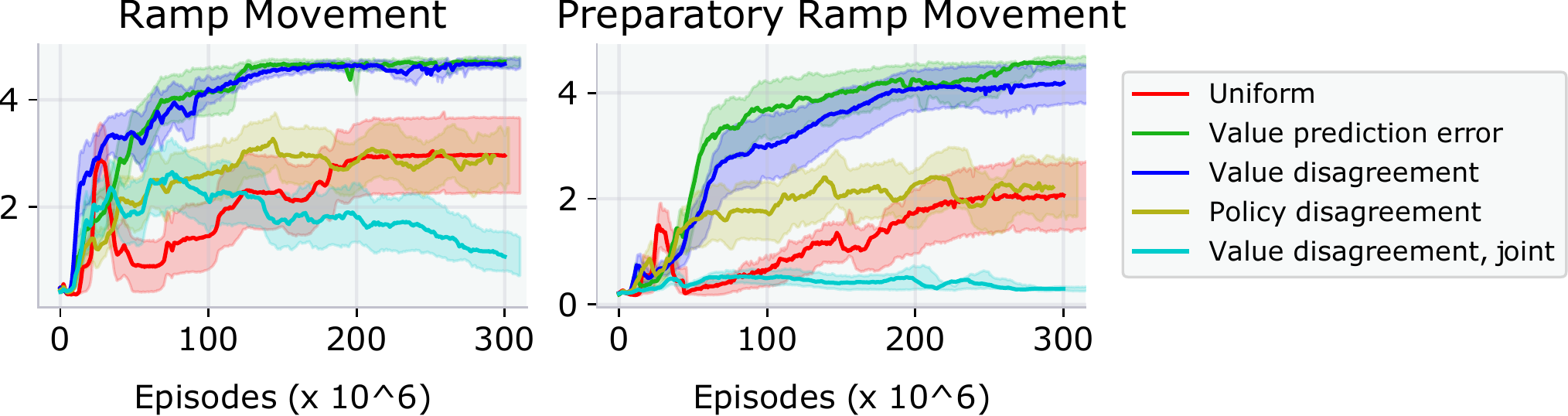}
    \caption{Ramp movement and preparatory ramp movement in Hide and Seek for different teacher rewards under uniform evaluation. In the ``Value disagreement, joint'' condition, the teacher controls agent, box and ramp spawn locations, in all other teacher conditions the teacher only controls box and ramp spawn locations. Shaded regions denote the standard error of the mean calculated from at least 8 random seeds.}
    \label{fig:hs_ramp-movement}
\end{figure}

We can therefore evaluate skill emergence of agents under different teacher rewards by monitoring ramp movement. Due to substantial seed dependence we report results averaged over at least 8 seeds in each condition. Training with value prediction error and value disagreement teacher reward leads to significantly larger ramp movement and preparatory ramp movement than under uniform sampling or with policy disagreement teacher reward (Figure \ref{fig:hs_ramp-movement}). The (preparatory) ramp movement averaged over seeds is lower in these conditions because phase 3 and 4 is only reached in a fraction of seeds. Phase 3 is reached for all value prediction error and value disagreement seeds, but only for 60\% uniform and policy disagreement seeds. Phase 4 is reached in 100 \% of value prediction error seeds, 89\% of value disagreement seeds, 36\% of policy disagreement seeds and 25\% of uniform seeds. Training with value prediction error and value disagreement teacher reward leads therefore to a much more robust skill discovery than under the uniform baseline and with policy disagreement teacher reward. A potential explanation for the discrepancy in performance between value and policy disagreement is that value disagreement measures epistemic uncertainty in students' value functions whereas policy disagreement measures epistemic uncertainty in the students' action distribution. Anecdotally, we observe that the emergence of a new skill in Hide and Seek corresponds to a rapid shift in value in most states, whereas the students' action distribution often only shifts substantially during key decision points (e.g. when the hider or seeker picks up a ramp). The larger shift in value disagreement than in policy disagreement might therefore make it a more reliable measure to detect environments that are conducive for skill discovery.

We also find that the default conditional sampling scheme where the teacher samples ramp and box spawn locations conditioned on random agent spawn locations performs much better than a joint sampling scheme where the teacher samples agent, ramp and box spawn locations (compare ``Value disagreement'' with ``Value disagreement, joint'' condition in Figure \ref{fig:hs_ramp-movement}). We speculate that conditioning the teacher on the distribution of uniformly spawned agents allows the teacher to better cover the distribution of currently learnable environments. 

\subsection{Doorless random maze with ramps}
\label{sec:random_maze}

In addition to multi-agent Hide and Seek, we also evaluate the teacher rewards in a single-agent random maze environment. Single-agent environments make it easier to analyze how the teacher shifts its sampling distribution as a function of student learning and to detect potentially pathological teacher behavior. We designed the random maze environment such that only a small fraction of environments are solvable for the agent, allowing us to verify that the teacher avoids the subspace of unsolvable environments.

In this environment, the agent needs to reach a goal that consists of a movable box in a procedurally generated random maze with 20 rooms. The agent receives a per-time step reward of +1 whenever it is close to the box. Since the rooms in the maze have no doors the agent can only move to a different room by using a movable ramps to climb a wall and balancing along narrow walls to the desired room. The setup of this environment (including physical properties of agent, box and ramp and the policy of teacher and student agent) is very similar to the hide and seek quadrant task. The only difference is that we disable line-of-sight and vision cone masking to allow the agent to locate the goal box even if it is hidden from sight. 

\begin{figure}
    \centering
    \includegraphics[width=\linewidth]{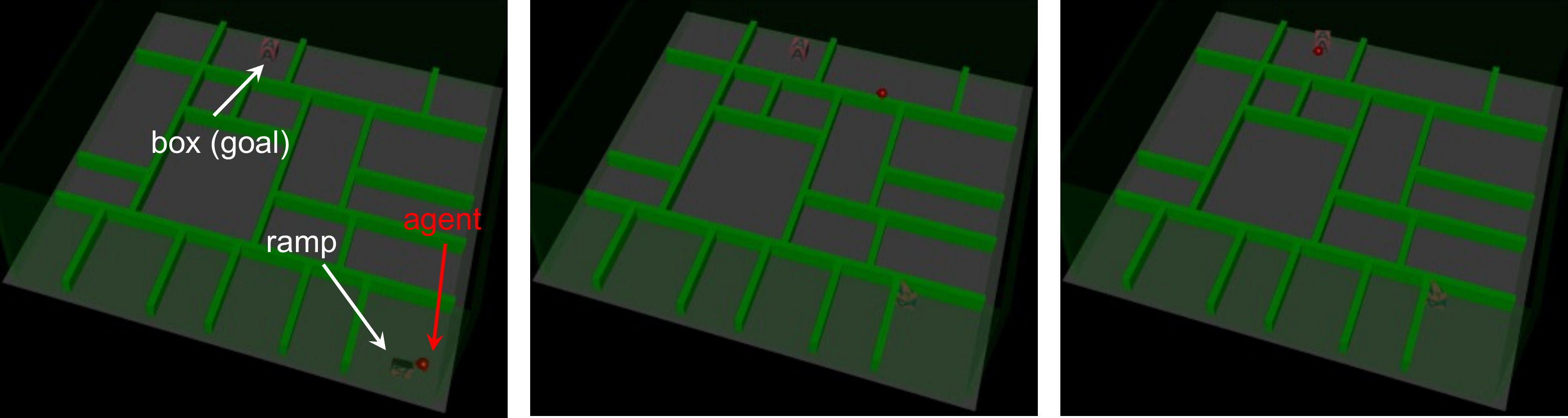}
    \caption{The agent solves a ``hard'' random maze environment: {\it Left:} The agent is spawned in the same room as the ramp, but in a different room than the box it needs to reach. {\it Middle}: The agent uses the ramp to climb a wall and balances over narrow walls to the room with the box. {\it Right}: The agent has successfully reached the box. }
    \label{fig:rm-example}
\end{figure}

Under uniform sampling, the agent, box and ramps are spawned randomly in the maze. Depending on their spawn locations, environments have 3 levels of difficulty:
\begin{itemize}
    \item {\it Easy}: The agent is spawned in the same room as the box and can maximize its reward simply by moving to the box and staying there for the remainder of the episode. About 7\% of uniformly sampled environments are in this category.
    \item {\it Hard}: The agent is spawned in the same room as the ramp, but in a different room than the box. The agent needs to use the ramp to climb the walls of the maze and balance on top of narrow walls to the room with the box (Figure \ref{fig:rm-example}). About 7\% of uniformly sampled environments are in this category.
    \item {\it Impossible}: The agent is spawned in a room without ramps or boxes. In this setting the agent cannot receive any reward. About 86\% of uniformly sampled environments are in this category.
\end{itemize}
 When training with teacher, the teacher observes the random maze and generates the grid locations of the agent, the box and the ramps. This corresponds to conditional sampling (see section \ref{sec:tscl}) where $Y$ parameterizes the random maze and $X$ parameterizes agent, box and ramp locations. We expect a successful teacher to sample easy environments early in training. Once the agent has discovered how to use the ramp to move over walls the teacher should predominantly sample hard environment. The teacher should avoid sampling impossible environments as much as possible.

 \begin{figure}
     \centering
     \includegraphics[width=\linewidth]{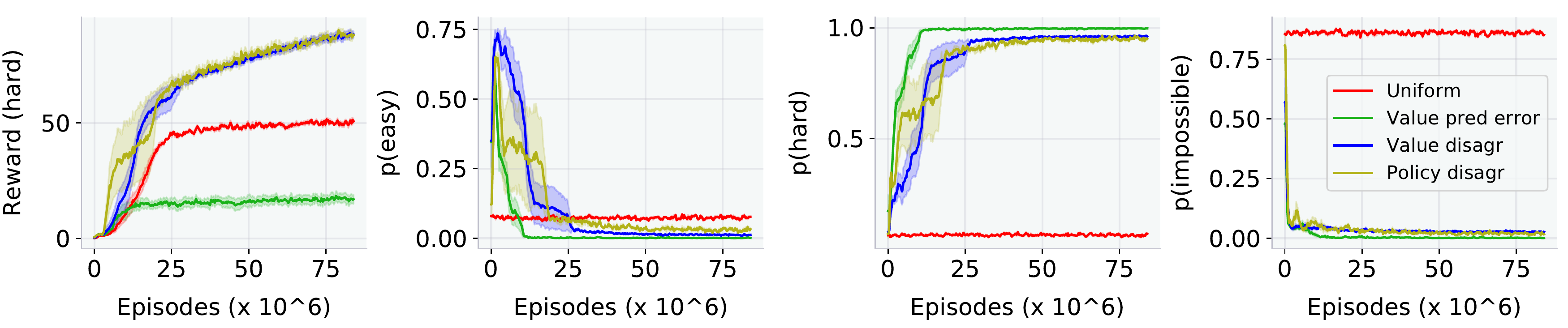}
     \caption{Agent performance and teacher behavior in the doorless random maze task: {\it Left:} Agent reward evaluated under uniformly sampled hard environments when trained with different teacher rewards. {\it 2nd, 3rd and 4th from left:} Probability of the teacher sampling easy, hard and impossible environments, respectively. Shaded regions correspond to standard error of the mean calculated from at least 6 seeds per condition. }
     \label{fig:rm_results}
 \end{figure}
 
 \begin{figure}
     \centering
     \includegraphics[width=\linewidth]{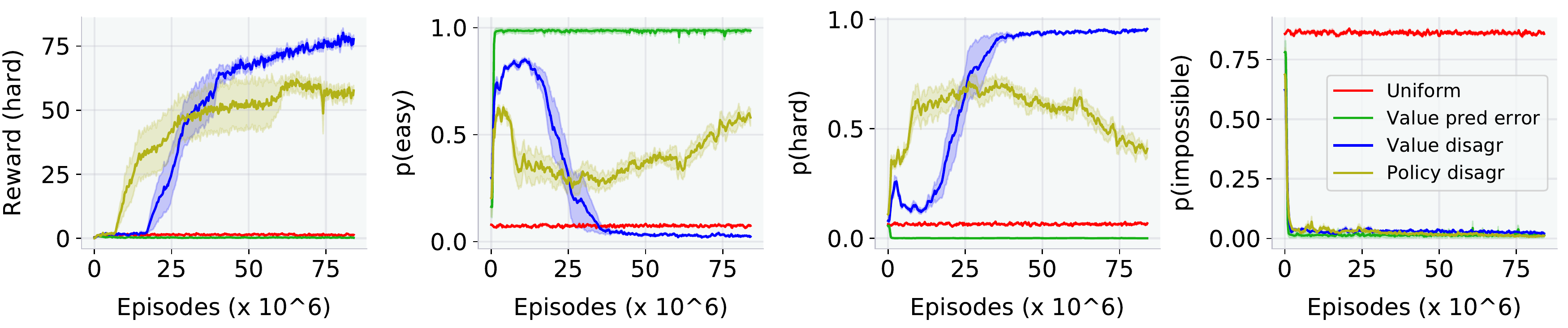}
     \caption{Agent performance and teacher behavior in the doorless random maze task with stochastic reward. Plots are analogous to Figure \ref{fig:rm_results}. }
     \label{fig:rm_results2}
 \end{figure}

We evaluate the agent's reward under uniformly sampled hard environments (i.e. we uniformly sample environments, but reject those that are not classified as hard). Training with value disagreement and policy disagreement teacher rewards leads to significantly better agent performances than the uniform baseline. However, training with value prediction error teacher reward leads to even worse performance than uniform baseline. This is surprising at first because the teacher shows sensible sampling behavior under all 3 teacher rewards, including for value prediction error: At beginning of training it predominantly samples easy environments (Figure \ref{fig:rm_results}, 2nd from left) and later shifts towards hard environments (Figure \ref{fig:rm_results}, 3rd from left). All teachers consistently suppress impossible environments (Figure \ref{fig:rm_results}, right).

The inferior performance of the value prediction error teacher suggests that while the teacher does identify hard environments as interesting it focuses on a subset of hard environments where the agent cannot learn new skills. One potential explanation for this effect could be that in some environments the sensitivity of the physics simulator to initial conditions may make it very hard to improve the prediction of the value function. The value prediction error teacher may therefore be incentivized to focus on the subset of hard environments that create intractable prediction problems rather than those where the agent can learn new behavior.

\subsection{Doorless random maze with ramps and stochastic reward}
\label{sec_maze_stoch}

We can test the hypothesis that the value prediction error is more susceptible to stochasticity in the environment by a simple variant of the random maze task: In this variant, we sample a Bernoulli variable with probability 0.5 at the beginning of an episode to determine whether the episode is rewarded or not. If the episode is rewarded, the student receives twice the reward it would have received in the deterministic variant. If the episode is unrewarded the student does not receive any reward no matter whether it reaches the goal. The expected reward for any trajectory is therefore identical to the deterministic version of the task. But in the stochastic version, the student cannot arbitrarily reduce its value prediction error through learning, since the Bernoulli variable that determines whether an episode is rewarded is unpredictable. 

Our results indeed show that value prediction error teacher is much more affected by stochastic returns than the value disagreement and policy disagreement teacher (Figure \ref{fig:rm_results2}, left). The value prediction error teacher exclusively samples easy environments throughout training (Figure \ref{fig:rm_results2}, 2nd from left). The reason is as follows: In easy environments, the agent spends the most time next to the goal box, because the agent does not need to overcome any obstacles to reach the goal. Each time step next to the goal box generates an irreducible contribution to the value prediction error, because the agent cannot predict whether the episode is rewarded or not. Easy environments therefore generate a larger value prediction error reward than hard environments, which incentivizes the value prediction error teacher to exclusively sample easy environments. The focus of this teacher on easy environments is analogous to the ``Noisy-TV'' phenomenon, where an agent that is rewarded for seeking unpredictable experience is attracted by unpredictable random noise \cite{burda2018large}.

Conversely, the value disagreement and policy disagreement teacher reach a performance in hard environments that is comparable to the deterministic task (Figure \ref{fig:rm_results2}, left). This is particularly remarkable because the stochasticity of the reward makes the task overall harder to learn (unlike in the deterministic task the agent does not learn any hard environments under uniform sampling). As in the deterministic task, the value disagreement teacher first focuses on easy environments and then shifts its sampling distribution to hard environments once the agent has started learning how to use the ramp. The policy disagreement teacher displays a similar pattern, but the shift is more gradual. Curiously, the policy disagreement teacher increases its sampling of easy environments late in training, which suggests that it is somewhat susceptible to stochasticity, but much less so than the value disagreement teacher.

\section{Discussion}
\label{sec:discussion}
We studied the problem of automatic environment design for multi-agent environments using an RL-trained teacher that samples environments to maximize student learning. As student reward is often a poor correlate of student performance in multi-agent settings, we focused on intrinsic teacher rewards that measure the discrepancy between a student reward or behavior with a prediction of the same. We found that value prediction error and value disagreement lead to faster and more reliable skill discovery in multi-agent hide and seek than uniform sampling or policy disagreement. However, value prediction error performed poorly in our diagnostic single-agent maze task, particularly if we made the student reward stochastic, while value disagreement was much more robust. These results suggest value disagreement as a promising teacher reward for automatic environment design for both multi-agent environments and environments with stochasticity.

While the main focus of this work was to identify teacher rewards for multi-agent environments, intrinsic teacher rewards such as value disagreement might also be attractive candidates for single agent settings. In particular, we have found that many previously proposed teacher reward schemes fall prey to adversarial situations where the teacher reward can be decoupled from genuine student learning progress. As an example, rewarding the teacher if the student's reward is in some set interval [a,b] could be gamed in situations where the maximal reward varies by environment. As another example, a scheme where the teacher is rewarded when the probability of success of the student lies in some interval [a,b], can be gamed if the maximum expected probability of success varies by environment. Expanding on latter example, if the level of noise in the environment is under the teacher's control, then the teacher can simply increase the noise until the success probability of the optimal policy is within the desired range.

Our results indicate that by sampling learnable environments, a well-designed teacher can speed up significantly the exploration of the student. However, a limitation of our approach is that it might not be applicable to environments where student rewards under uniform sampling are very sparse, because in the total absence of any student reward intrinsic teacher rewards vanish as well. Furthermore, while the domain independence of intrinsic teacher rewards make them easily adaptable in new settings with little hyperparameter tuning, they might cause the emergence of student skills that are not interesting to humans or that are not extractable via fine-tuning. Finally, intrinsic teacher rewards such as value or policy disagreement require training of a second value or policy network, which might be costly for large networks.



{
\small

\bibliographystyle{unsrt}
\bibliography{bibliography.bib}

}


\appendix

\section{Appendix}

\subsection{AutoDIME algorithm}
\label{app_dime}

\begin{algorithm}[H]
\SetAlgoLined
Randomly initialize teacher policy $\pi^T$ and student policy $\pi^S$\;
Split environment parameters into $Z=(X,Y)$, where $Y\sim p(Y)$ is sampled from a stationary distribution and $X|Y \sim \pi^T(Y)$ is sampled from the teacher policy\;
\While{not converged}{
    Sample stationary environment parameters: $Y \sim p(Y)$\;
    Sample remaining parameters using teacher policy: $X|Y \sim \pi^T(Y)$\;
    Roll out students in environment $Z=(X,Y)$\ and collect their rewards\;
    Compute per-time step teacher reward $r_t$ using one of the teacher rewards in section \ref{sec:teacher_rewards}\;
    Sum per-time step teacher reward over episode time and average over (on-policy) students\;
    Train student policy $\pi^S$ with RL update and student rewards\;
    Train teacher policy $\pi^T$ with RL update and teacher reward\;
}
\caption{AutoDIME}
\end{algorithm}

\section{Policy architectures}
\label{app_policy}
\subsection{Student policy architecture}
The student policy architecture was very similar to \cite{baker2020emergent}: Agents' observations of episode time, their own locations and velocities, the location and velocities of other agents, and the location and velocities of boxes and ramps were each embedded in a 256-dimensional space. Objects of the same type (i.e. other agents, boxes or ramps) were embedded using the same weights with entity-centric observations \cite{dvzeroski2001relational, diuk2008object}. The LIDAR-like observation (which measured the distance to the nearest objects using 30 evenly arrayed radial range sensors) was passed through a 1d-convolutional layer with 9 filters and kernel size 3 before being embedded together with agents' observations of their own locations and velocities. We then pooled over entities using mean-pooling. In the Hide and Seek environment, invisible agents or objects where masked out before mean-pooling. Further processing after the pooling layer consisted of a fully-connected layer, an LSTM (each of size 256), a layer norm and a factorial action head. Each value function consisted of a separate network whose architecture is identical to the policy network. Masks where not used for value functions (i.e. the value function observed both visible and invisible objects).

\subsection{Teacher policy architecture}
The teacher observed a 30x30 occupancy grid of walls and objects that were sampled using the stationary distribution $X \sim p(X)$. The teacher observation was processed through the 18-layer residual convolutional IMPALA network \cite{espeholt2018impala}, followed by a fully connected layer of size 256, a layer norm and a factorial action head that specified the x and y grid position of each object the teacher controls. The value function used a separate network of identical architecture.

\section{Optimization hyperparameters}
\label{app_hyper}
For the training of the student policy we used hyperparameters similar to \cite{baker2020emergent}, except a 4 times smaller buffer size and mini-batch size (to limit the available compute). Buffer and mini-batch size for the teacher was chosen such that the time of each optimization iteration was similar for teacher and student policy. The student and teacher policy was trained on four and one A100 NVIDIA GPUs, respectively. The training time was 8 days for the Hide and Seek task and 55 hours for the maze tasks.

\begin{table}[h!]
\centering
\begin{tabular}{|c|c|}
    \hline
    Buffer size (student) & 80,000 \\
    \hline
    Buffer size (teacher, Hide and Seek) & 1250 \\
    \hline 
    Buffer size (teacher, maze tasks) & 5000 \\
    \hline
    Mini-batch size (student) & 16,000 x 10 time steps \\
    \hline
    Mini-batch size (teacher, Hide and Seek) & 250 x 1 time step \\
    \hline
    Mini-batch size (teacher, maze tasks) & 1000 x 1 time step \\
    \hline
    Learning rate & $3 \cdot 10^{-4}$ \\
    \hline
    PPO clipping parameter & 0.2 \\
    \hline
    Entropy coefficient & 0.01 \\
    \hline
    $\gamma$ & 0.998 \\
    \hline
    GAE parameter $\lambda$ & 0.95 \\
    \hline
    BPTT truncation length (student) & 10 \\
    \hline
    student episode length & 160 \\
    \hline
    teacher episode length & 1 \\
    \hline
\end{tabular}
\end{table}

\end{document}